\documentclass{article}

\usepackage{microtype}
\usepackage{graphicx}
\usepackage{subcaption}
\usepackage{booktabs} 
\usepackage{lingmacros}
\usepackage{tree-dvips}
\usepackage{multicol}
\usepackage{gensymb}        
\usepackage{graphicx}
\usepackage{amsmath}
\usepackage{amssymb}
\usepackage{epstopdf}
\usepackage{threeparttable}
\usepackage{multirow}
\usepackage{soul}
\usepackage{xspace}

\makeatletter
\DeclareRobustCommand\onedot{\futurelet\@let@token\@onedot}
\def\@onedot{\ifx\@let@token.\else.\null\fi\xspace}

\def\eg{\emph{e.g}\onedot} 
\def\ie{\emph{i.e}\onedot}

\def\etal{\emph{et al}\onedot}
\makeatother

\newcommand{\remotenet}{ReMotENet}

\usepackage{hyperref}
\usepackage{cleveref}

\usepackage[accepted]{icml2019}

\icmltitlerunning{}

\begin{document}

\twocolumn[
\icmltitle{Movement Tracks for the Automatic Detection of Fish Behavior in Videos}



\icmlsetsymbol{equal}{*}

\begin{icmlauthorlist}
\icmlauthor{Declan McIntosh}{ceng}
\icmlauthor{Tunai Porto Marques}{ceng}
\icmlauthor{Alexandra Branzan Albu}{ceng}
\icmlauthor{Rodney Rountree}{bio}
\icmlauthor{Fabio De Leo}{onc}
\end{icmlauthorlist}

\icmlaffiliation{bio}{Biology Department, University of Victoria, BC, Canada}
\icmlaffiliation{onc}{Ocean Networks Canada, BC, Canada}
\icmlaffiliation{ceng}{University of Victoria, BC, Canada}

\icmlcorrespondingauthor{Alexandra Branzan Albu}{aalbu@uvic.ca}

\icmlkeywords{Machine Learning, Climate Change, Deep Learning, Fish Detection, Event Detection, Video Processing}

\vskip 0.3in
]

\printAffiliationsAndNotice{} 





\begin{abstract}
Global warming is predicted to profoundly impact ocean ecosystems. Fish behavior is an important indicator of changes in such marine environments. Thus, the automatic identification of key fish behavior in videos represents a much needed tool for marine researchers, enabling them to study climate change-related phenomena. We offer a dataset of sablefish (Anoplopoma fimbria) startle behaviors in underwater videos, and investigate the use of deep learning (DL) methods for behavior detection on it. Our proposed detection system identifies fish instances using DL-based frameworks, determines trajectory tracks, derives novel behavior-specific features, and employs Long Short-Term Memory (LSTM) networks to identify startle behavior in sablefish. Its performance is studied by comparing it with a state-of-the-art DL-based video event detector.   
\end{abstract}

\section{Introduction}
Among the negative impacts of climate change in marine ecosystems predicted for global warming levels of $1.5$\degree C to $2$\degree C (\eg~significant global mean sea level rise~\cite{stocker2014climate}, sea-ice-free Artic Oceans~\cite{bindoff2013detection}, interruption of ocean-based services) are the acidification and temperature rise of waters. 


The behavioral disturbance in fish species resulting from climate change can be studied with the use of underwater optical systems, which have become increasingly prevalent over the last six decades \cite{mallet2014underwater,mdpipaper, ONC_Enviro}. However, advancements in automated video processing methodologies have not kept pace with advances in the video technology itself. The manual interpretation of visual data requires prohibitive amounts of time, highlighting the necessity of semi- and fully-automated methods for the enhancement \cite{porto2020l2uwe,ancuti2012enhancing} and annotation of marine imagery. \par

As a result, the field of automatic interpretation of underwater imagery for biological purposes has experienced a surge of activity in the last decade~\cite{aguzzi2020potential}. While numerous works propose the automatic detection and counting of specimen \cite{toh2009automated,spampinato2008detecting,zhang2020automatic}, ecological applications require more complex insights. Video data provides critical information on fish behavior and interactions such as predation events, aggressive interactions between individuals, activities related to reproduction and startle responses. 
The ability to detect such behavior represents an important shift in the semantic richness of data and scientific value of computer vision-based analysis of underwater videos: from the focused detection and counting of individual specimens, to the context-aware identification of fish behavior.\par


Given the heterogeneous visual appearance of diverse behaviors observed in fish, we initially focus our study on a particular target: startle motion patterns observed in sablefish (\textit{Anoplopoma fimbria}). Such behavior is characterized by sudden changes in the speed and trajectory of sablefish movement tracks. \par

We propose a novel end-to-end behavior detection framework that considers 4-second clips to 1) detect the presence of sablefish using DL-based object detectors~\cite{redmon2018yolov3}; 2) uses the Hungarian algorithm \cite{kuhn1955hungarian} to determine trajectory tracks between subsequent frames; 3) measures four handcrafted and behavior-specific features and 4) employs such features in conjunction with LSTM networks~\cite{hochreiter1997long} to determine the presence of startle behavior and describe it (\ie~travelling direction, speed, and trajectory). The remainder of this article is structured as follows. In \Cref{sec:prev} we discuss works of relevance to the proposed system. \Cref{sec:PA} describes the proposed approach. In \Cref{sec:Results_Disc} we present a dataset of sablefish startle behaviors and use it to compare the performance of our system with that of a state-of-the-art event detector~\cite{yu2018remotenet}. \Cref{sec:con} draws conclusions and outlines future work.         

\begin{figure*}[t]
\begin{center}
\includegraphics[width=1\linewidth]{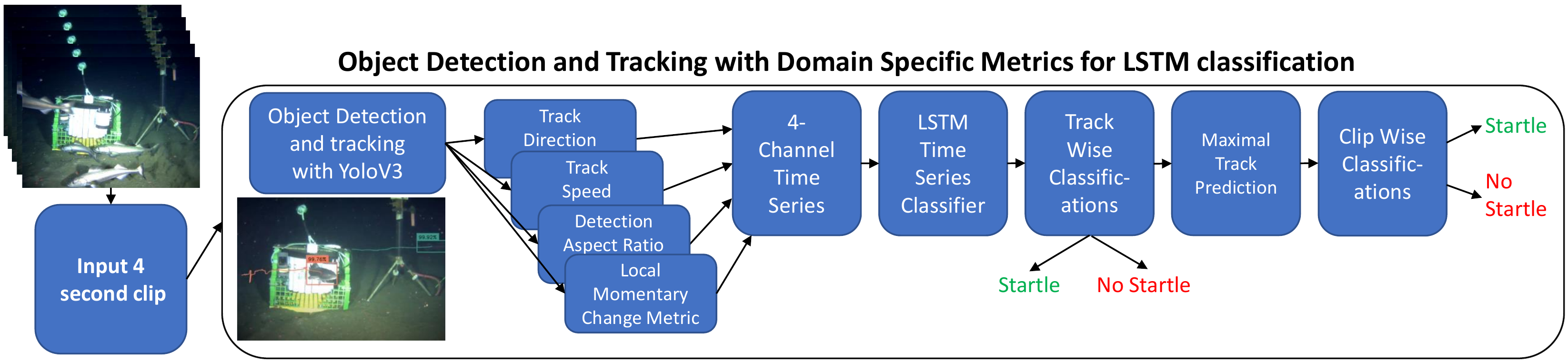}
\end{center}
   \caption{Computational pipeline of the fish behavior detection system proposed. The framework is able to provide clip-wise and movement track-wise classifications alike (see \ref{sec:ex_results}).}
\label{fig:1-overall}
\end{figure*}

\section{Related Works}
\label{sec:prev}

Related works to our approach include DL-based methods for object detection in images and events in videos. 

\textbf{Deep learning-based object detection for static images}. Krizhevsky \etal~\cite{krizhevsky2012imagenet} demonstrated the potential of using Convolutional Neural Networks (CNNs) to extract and classify visual features from large datasets. Their work motivated the use of CNNs in object detection, where frameworks perform both \textit{localization} and \textit{classification} of regions of interest (RoI). Girshick \etal~\cite{girshick2014rich} introduced R-CNN, a system that uses a traditional computer vision-based technique (selective search \cite{uijlings2013selective}) to derive RoIs where individual classification tasks take place. Processing times are further reduced in Fast R-CNN \cite{girshick2015fast} and Faster R-CNN~\cite{ren2015faster}. A group of frameworks \cite{liu2016ssd, redmon2016you, lin2017focal} referred to as ``one-stage'' detectors proposed the use of a single trainable network for both creating RoIs and performing classification. This reduces processing times, but often leads to a drop in accuracy when compared to two-stage detectors. Recent advancements in one-stage object detectors (\eg a loss measure that accounts for extreme class imbalance in training sets \cite{lin2017focal}) have resulted in frameworks such as YOLOv3~\cite{redmon2018yolov3} and RetinaNet~\cite{lin2017focal}, which offer fast inference times and performances comparable with that of two-stage detectors.\par

\textbf{DL-based event detection in videos}. Although object detectors such as YOLOv3~\cite{redmon2018yolov3} can be used in each frame of a video, 
they often ignore important temporal relationships. Rather than individual images employed by aforementioned methods, recent works \cite{kang2017noscope,yu2018remotenet,saha2016deep,xu2015learning,ionescu2019object,cocsar2016toward,xu2019joint} used video's inter-frame temporal relationship to detect relevant events. Saha \etal~\cite{saha2016deep} use Fast R-CNN \cite{girshick2015fast} to identify motion from RGB and optical flow inputs. The outputs from these networks are fused resulting in action tubes that encompass the temporal length of each action. 
Kang \etal~\cite{kang2017noscope} offered a video querying system that trains specialized models out of larger and more general CNNs to be able to efficiently recognize only specific visual targets under constrained view perspectives with claimed processing speed-ups of up to $340\times$. Co{\c{s}}ar \etal~\cite{cocsar2016toward} combined an object-tracking detector \cite{chau2011robust}, trajectory- and pixel- based methods to detect abnormal activities. Ionescu \etal~\cite{ionescu2019object} offered a system that not only recognizes motion in videos, but also considers context to differentiate between normal (\eg a truck driving on a road) and abnormal (\eg a truck driving on a pedestrian lane) events. \par

Yu \etal\cite{yu2018remotenet} proposed \remotenet, a light-weight event detector that leverages spatial-temporal relationships between objects in adjacent frames. It uses 3D CNNs (``spatial-temporal attention modules'') to jointly model these video characteristics using the same trainable network. A frame differencing process allows for the network to focus exclusively on relevant, motion-triggered regions of the input frames. This simple yet effective architecture results in fast processing speeds and reduced model sizes \cite{yu2018remotenet}. \par
 
\section{Proposed approach}
\label{sec:PA}

We propose a hybrid method for the automatic detection of context-aware, ecologically relevant behavior in sablefish. We first describe our method for tracking sablefish in video, then propose the use of 4 track-wise features to constrain sablefish startle behaviors. Finally, we describe a Long Short Term Memory (LSTM) architecture that performs classification using the aforementioned features. 


\textbf{Object detection and tracking}. We use the YOLOv3 \cite{redmon2018yolov3} end-to-end object detector as the first step of this hybrid method. The detector was completely re-trained to perform a simplified detection task were only the class \textit{fish} is targeted. We use a novel 600-image dataset (detailed in \ref{sec:dataset}) of sablefish instances composed of data from Ocean Networks Canada (ONC) to train the object detector. 


The detection of each frame offer a set of bounding boxes and spatial centers. In order to track organisms we associate these detection between frames. Our association loss value consists simply of the distance between detection centers in two subsequent frames. We employ the Hungarian Algorithm \cite{kuhn1955hungarian} to generate a loss minimizing associations between detection in two consecutive frames. We then remove any associations where the distance between various detection is greater than 15\% of the frame resolution. Tracks are terminated if no new detection is associated with them for 5 frames (\ie 0.5 seconds---see \ref{sec:dataset}).


\begin{figure}[h]
\centering
\includegraphics[width=1\linewidth]{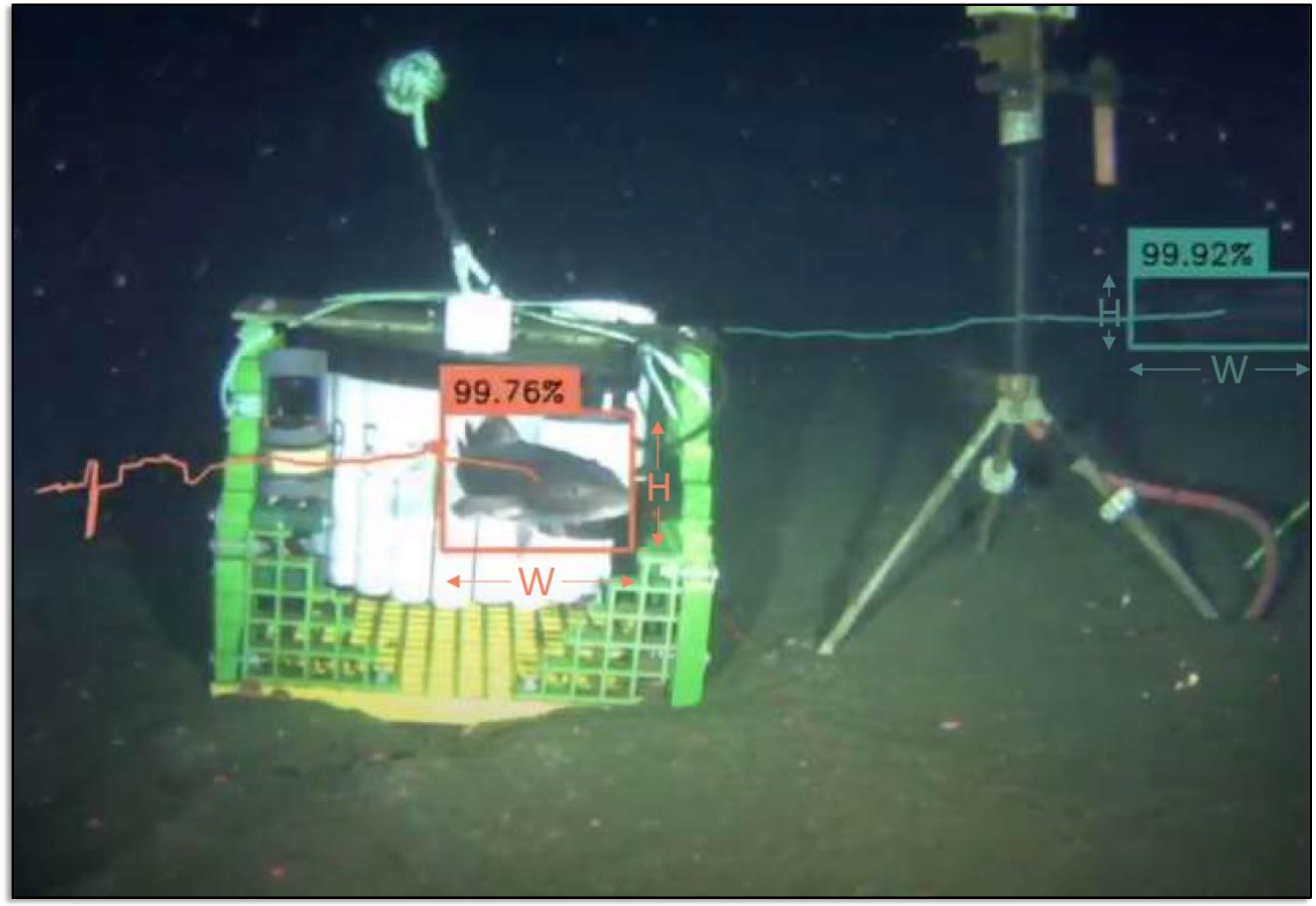}
\vspace{-1.7em}
\caption{Sample movement tracks and object detection confidence scores. The bounding boxes highlight the \textit{fish} detection in the current frame. Each color represents an individual track.}
\end{figure}

\textbf{Behavior Specific Features}. We propose the use of a series of four domain-specific features that describe the startle behavior of sablefish. Each feature conveys independent and complementary information, and the limited number of features ($4$) prevents over-constraining the behavior detection problem. \par

The first two features quantify the \textit{direction of travel} and \textit{speed} from a track. These track characteristics were selected because often the direction of travel changes and the fish accelerates at the moment of a startle motion. \par

\begin{figure}[h]
\centering
\includegraphics[width=0.5\linewidth]{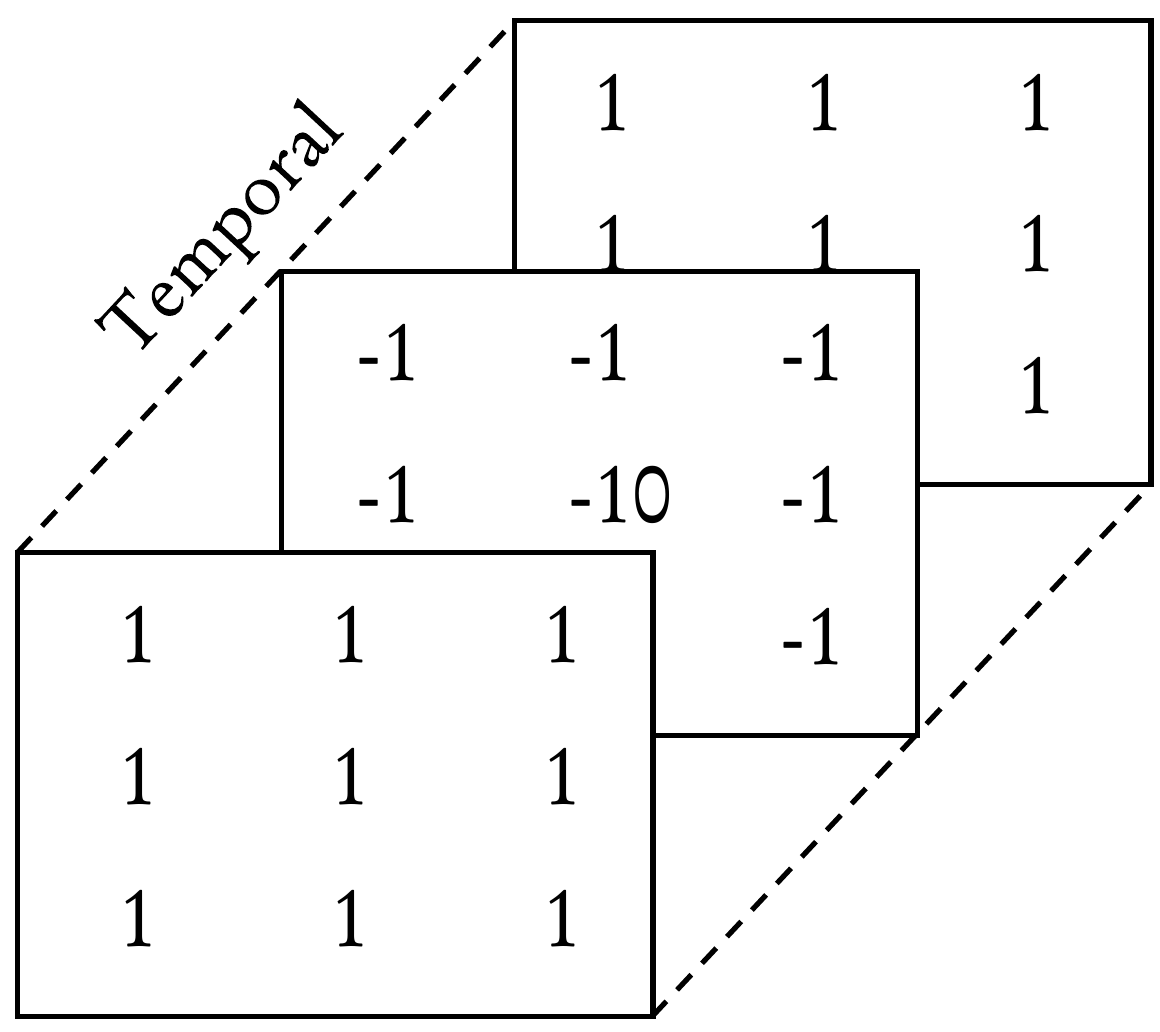}
\vspace{-0.5em}
\caption{LMCM kernel designed to extract fast changes in sequential images.}
\label{fig:LOG-Kernel}
\end{figure}

A third metric considers the \textit{aspect ratio} of the detection bounding box of a \textit{fish} instance over time. The reasoning behind this feature is the empirical observation that sablefish generally take on a ``c'' shape when startling, in preparation for moving away from their current location. The final \textit{Local Momentary Change Metric (LMCM)} feature seeks to find fast and unstained motion, or temporal visual impulses,  associated with startle events. This feature  is obtained by convolving the novel 3-dimensional LMCM  kernel, depicted in \Cref{fig:LOG-Kernel}, over three temporally adjacent frames. This spatially symmetric kernel was designed to produce high output values where impulse changes occur between frames. Given its zero-sum and isotropic properties, the kernel outputs zero when none or only constant temporal changes are occurring. We observe that the LMCM kernel efficiently detects leading and lagging edges of motion. In order to associate this feature with a track we average the LMCM output magnitude inside a region encompassed by a given fish detection bounding box.


\subsection{LSTM classifier}

\begin{figure}[h]
\centering
\includegraphics[width=1\linewidth]{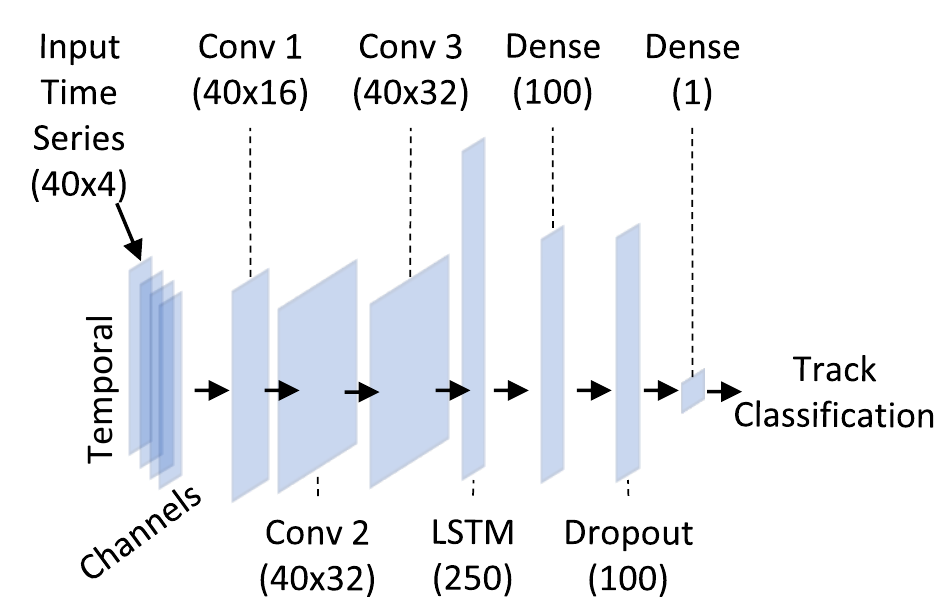}
\vspace{-0.5em}
\caption{LSTM-based network for the classification of movement tracks. The network receives four features conveying speed, direction and rate of change from each track as input.}
\label{fig:LSTM Classifier}
\end{figure}

In order to classify an individual track, we first combine its four features as a \textit{tensor} data structure of dimensions $(40,4)$; tensors associated with tracks of less than $40$ frames are end-padded with zeros. A set of normalization coefficients calculated empirically using the training set (see~\ref{sec:dataset}) is then used to normalize each value in the input time-series to the range $[-1,1]$. A custom-trained long short-term memory (LSTM) classifier receives the normalized tensors as input, and outputs a track classification of \textit{non-startle background} or \textit{startle}, as well as a confidence score. This is done by considering underlying temporal relationships between their values. We chose to use LSTM networks because the temporal progression of values from the features extracted (along 40 frames or 4 seconds) conveys important information for the classification of individual clips/tracks. \Cref{fig:LSTM Classifier} details the architecture of the LSTM network employed. All convolutional layers employ three-layered 1D kernels.\par 

\section{Results and Discussion}
\label{sec:Results_Disc}

We compare our method with a state-of-the-art event detection algorithm \cite{yu2018remotenet}. Section \Cref{sec:dataset} describes our dataset.  Our comparison considers standard performance metrics in \Cref{sec:ex_results} and discusses the potential advantages of semantically richer data outputs in ecological research.   

\subsection{Sablefish Startle Dataset}
\label{sec:dataset}


The data used in this work was acquired with a stationary video camera permanently deployed at $620$m of depth at the Barkley Node of Ocean Networks Canada's~\footnote{www.oceannetworks.ca/} NEPTUNE marine cabled observatory infrastructure. All videos samples were collected between September 17\textsuperscript{th} and October 24\textsuperscript{th} 2019 because this temporal window contains high sablefish activity. The original  monitoring videos are first divided into units of 4-second clips (a sablefish startle is expected to last roughly one second) and down-sampled to 10 frames per second for processing reasons. An initial filtering is carried out using Gaussian Mixture Models~\cite{stauffer1999adaptive}, resulting in a set composed only by clips which contain motion. For training purposes, these motion clips are then manually classified as possessing startle or not. The Sablefish Startle dataset consists of $446$ positive (\ie presenting startle events) clips, as  well as 446 randomly selected negative samples (\ie without startle events). \Cref{tab:2-data-split} details the dataset usage for training, validation and testing.\par        

\begin{table}[h]
\scriptsize
\begin{center}
\begin{tabular}{p{0.15  \linewidth}  p{0.1\linewidth}  p{0.18\linewidth}  p{0.1\linewidth}  p{0.18\linewidth}}
\hline
\textbf{Data Split}  & \textbf{Clips} &  \textbf{Startle Clips} & \textbf{Tracks} & \textbf{Startle Tracks}\\
\hline

Train	        & 642 & 321	&	 1533	       & 323 \\
Validation      & 150 & 75	&	 421	       & 80 \\
Test            & 100  & 50	&	 286	       & 50 \\

\hline
\end{tabular}
\end{center}
\vspace{-1.4em}
\caption{Division of the 4-second clips of the Sablefish Startle Dataset for training, validation and testing purposes.}
\label{tab:2-data-split}
\end{table}
\vspace{-0.6em}


A second dataset composed of 600 images of sabblefish was created to train the YOLOV3~\cite{redmon2018yolov3} object detector (\ie first step of the proposed approach). In order to assess the track-creation performance of the proposed system, we use this custom-trained object detector to derive movement tracks from each of the 892 clips composing the Sablefish Startle Dataset. Tracks with length shorter than 2 seconds are discarded. The remaining tracks are manually annotated as startle or non-startle (see~\Cref{tab:2-data-split}). 

This dual annotation approach (\ie clip- and track-wise) employed with the Sablefish Startle Dataset allows for a two-fold performance analysis: 1) \textit{clip-wise classification}, where an entire clip is categorized as possessing startle or not, and 2) \textit{track-wise classification}, which reflects the accuracy in the classification of each candidate track as startle or non-startle.   


\subsection{Experimental Results}
\label{sec:ex_results}

%

We calculate the Average Precision (AP), Binary Cross Entropy (BCE) loss and Recall for both track- and clip-wise outputs of the proposed system using only the Test portion of the Sablefish Startle Dataset. A threshold of $0.5$ (in a $[0,1]$ range) is set to classify a candidate movement track as positive or negative with respect to all of its constituent points. In order to measure the performance of the clip-wise classification and compare it with the baseline method (\remotenet~\cite{yu2018remotenet}), we consider that the ``detection score'' of a clip is that of its highest-confidence movement track (if any). Thus, any clip where at least one positive startle movement track is identified will be classified as positive.      




The conversion from track-wise classification to clip-wise classification is expected to lower the overall accuracy of our proposed approach. A ``true'' startle event might create only short, invalid tracks, or sometimes no tracks at all. This situation would lower the clip-wise classification performance, but would not interfere with the track-wise one. The track-wise metrics are applicable only to our approach and they mainly reflect the difference between the manual and automatic classification of the tracks created in the dataset by the proposed system, thus evaluating the ability of the LSTM network to classify tracks. Table~\ref{tab:2-results} shows that the LSTM portion of our method performs well for classifying startle tracks (AP of $0.85$). Clip-wise, our method outperformed a state-of-the-art DL-based event detector~\cite{yu2018remotenet} with an AP of $0.67$. 


\begin{table}[h]
\scriptsize
\begin{center}
\begin{tabular}{p{0.19  \linewidth}  p{0.13  \linewidth}  p{0.15\linewidth}  p{0.13\linewidth}  p{0.14\linewidth}}
\hline
\textbf{Method}  & \textbf{Track AP} & \textbf{Track BCE} & \textbf{Clip AP} & \textbf{Clip Recall}\\
\hline

Ours	                                & \textbf{0.85} & \textbf{0.412}	&	 \textbf{0.67}	    & \textbf{0.58}  \\
\remotenet~\cite{yu2018remotenet}	    & N/A$^{1}$  & N/A$^{1}$	&	 0.61	       & 0.5 \\
\hline
\end{tabular}
\end{center}
\begin{tablenotes}
\item[1] $^{1}$: \remotenet does not perform track-wise classification.       
\end{tablenotes}
\vspace{-1.2em}
\caption{Performance comparison between the proposed method and a state-of-the-art event detector.}
\label{tab:2-results}
\end{table}
\vspace{-0.75em}

Our proposed system generates more semantically rich data by detecting and describing behavior rather than just marking a clip as containing the behavior; this may be helpful for ecological and biological research. Our approach provides instance-level information such as track-specific average speed, direction and rate of change. These extra data may allow for further analyses considering inter- and intra-species behaviors. Also, there are clips that contain more than one startle, and our approach is able to identify all startle instances in such clips. Simply classifying a clip as startle or non-startle would, of course, not allow for the detection of multiple startle instances within the same clip.  


\section{Conclusion}
\label{sec:con}
We propose an automatic detector of fish behavior in videos that performs semantically richer tasks than typical computer vision-based analyses: instead of specimens counting, it identifies and describes a complex biological event (startle events of sablefish). Our intent is to enable long-term studies on changes in fish behavior that could be caused by climate change (\eg temperature rise and acidificaton). 


A dataset composed of 892 4-second positive (startle) and negative (non-startle) clips, and associated tracks were manually annotated. Experiments using this data show that the proposed detector identifies and classifies well individual tracks of motion as startle or not (AP of $0.85$). Furthermore, the performance of our clip-wise classification is compared to that of a state-of-the-art event detector, \remotenet~\cite{yu2018remotenet}. Our system outperforms \remotenet~with an AP of $0.67$ (against $0.61$). \par

Future work will address more fish behaviors (\eg predation, spawining) and will adapt DL-based event detectors such as \remotenet~\cite{yu2018remotenet} and NoScope~\cite{kang2017noscope} to that end. 


\bibliography{SHdetection}

\begin{thebibliography}{10}

\bibitem{stocker2014climate}
Thomas~F Stocker, Dahe Qin, G-K Plattner, Melinda~MB Tignor, Simon~K Allen,
  Judith Boschung, Alexander Nauels, Yu~Xia, Vincent Bex, and Pauline~M
  Midgley.
\newblock Climate change 2013: The physical science basis. contribution of
  working group i to the fifth assessment report of ipcc the intergovernmental
  panel on climate change, 2014.

\bibitem{bindoff2013detection}
Nathaniel~L Bindoff, Peter~A Stott, Krishna~Mirle AchutaRao, Myles~R Allen,
  Nathan Gillett, David Gutzler, Kabumbwe Hansingo, G~Hegerl, Yongyun Hu, Suman
  Jain, et~al.
\newblock Detection and attribution of climate change: from global to regional.
\newblock 2013.

\bibitem{mallet2014underwater}
Delphine Mallet and Dominique Pelletier.
\newblock Underwater video techniques for observing coastal marine
  biodiversity: a review of sixty years of publications (1952--2012).
\newblock {\em Fisheries Research}, 154:44--62, 2014.

\bibitem{mdpipaper}
Tunai Porto~Marques, Alexandra Branzan~Albu, and Maia Hoeberechts.
\newblock A contrast-guided approach for the enhancement of low-lighting
  underwater images.
\newblock {\em Journal of Imaging}, 5(10):79, 2019.

\bibitem{ONC_Enviro}
Jacopo Aguzzi, Carolina Doya, Samuele Tecchio, Fabio De~Leo, Ernesto Azzurro,
  Cynthia Costa, Valerio Sbragaglia, Joaquin del Rio, Joan Navarro, Henry Ruhl,
  Paolo Favali, Autun Purser, Laurenz Thomsen, and Ignacio Catalán.
\newblock Coastal observatories for monitoring of fish behaviour and their
  responses to environmental changes.
\newblock {\em Reviews in Fish Biology and Fisheries}, 25:463--483, 2015.

\bibitem{porto2020l2uwe}
Tunai Porto~Marques and Alexandra Branzan~Albu.
\newblock L2uwe: A framework for the efficient enhancement of low-light
  underwater images using local contrast and multi-scale fusion.
\newblock In {\em Proceedings of the IEEE/CVF Conference on Computer Vision and
  Pattern Recognition Workshops}, pages 538--539, 2020.

\bibitem{ancuti2012enhancing}
Cosmin Ancuti, Codruta~Orniana Ancuti, Tom Haber, and Philippe Bekaert.
\newblock Enhancing underwater images and videos by fusion.
\newblock In {\em 2012 IEEE Conference on Computer Vision and Pattern
  Recognition}, pages 81--88. IEEE, 2012.

\bibitem{aguzzi2020potential}
J.~Aguzzi1, D.~Chatzievangelou, J.B. Company, L.~Thomsen, S.~Marini,
  F.~Bonofiglio, F.~Juanes, R.~Rountree, A.~Berry, R.~Chumbinho, C.~Lordan,
  J.~Doyle, J.~del Rio, J.~Navarro, F.C. De~Leo, N.~Bahamon, J.A. García,
  R.~Danovaro, M.~Francescangeli, V.~Lopez-Vazquez1, and Ps~Gaughan.
\newblock The potential of video imagery from worldwide cabled observatory
  networks to provide information supporting fish-stock and biodiversity
  assessment.
\newblock {\em ICES Journal of Marine Science}, In press.

\bibitem{toh2009automated}
YH~Toh, TM~Ng, and BK~Liew.
\newblock Automated fish counting using image processing.
\newblock In {\em 2009 International Conference on Computational Intelligence
  and Software Engineering}, pages 1--5. IEEE, 2009.

\bibitem{spampinato2008detecting}
Concetto Spampinato, Yun-Heh Chen-Burger, Gayathri Nadarajan, and Robert~B
  Fisher.
\newblock Detecting, tracking and counting fish in low quality unconstrained
  underwater videos.
\newblock {\em VISAPP (2)}, 2008(514-519):1, 2008.

\bibitem{zhang2020automatic}
Song Zhang, Xinting Yang, Yizhong Wang, Zhenxi Zhao, Jintao Liu, Yang Liu,
  Chuanheng Sun, and Chao Zhou.
\newblock Automatic fish population counting by machine vision and a hybrid
  deep neural network model.
\newblock {\em Animals}, 10(2):364, 2020.

\bibitem{redmon2018yolov3}
Joseph Redmon and Ali Farhadi.
\newblock Yolov3: An incremental improvement.
\newblock {\em arXiv preprint arXiv:1804.02767}, 2018.

\bibitem{kuhn1955hungarian}
Harold~W Kuhn.
\newblock The hungarian method for the assignment problem.
\newblock {\em Naval research logistics quarterly}, 2(1-2):83--97, 1955.

\bibitem{hochreiter1997long}
Sepp Hochreiter and J{\"u}rgen Schmidhuber.
\newblock Long short-term memory.
\newblock {\em Neural computation}, 9(8):1735--1780, 1997.

\bibitem{yu2018remotenet}
Ruichi Yu, Hongcheng Wang, and Larry~S Davis.
\newblock Remotenet: Efficient relevant motion event detection for large-scale
  home surveillance videos.
\newblock In {\em 2018 IEEE Winter Conference on Applications of Computer
  Vision (WACV)}, pages 1642--1651. IEEE, 2018.

\bibitem{krizhevsky2012imagenet}
Alex Krizhevsky, Ilya Sutskever, and Geoffrey~E Hinton.
\newblock Imagenet classification with deep convolutional neural networks.
\newblock In {\em Advances in neural information processing systems}, pages
  1097--1105, 2012.

\bibitem{girshick2014rich}
Ross Girshick, Jeff Donahue, Trevor Darrell, and Jitendra Malik.
\newblock Rich feature hierarchies for accurate object detection and semantic
  segmentation.
\newblock In {\em Proceedings of the IEEE conference on computer vision and
  pattern recognition}, pages 580--587, 2014.

\bibitem{uijlings2013selective}
Jasper~RR Uijlings, Koen~EA Van De~Sande, Theo Gevers, and Arnold~WM Smeulders.
\newblock Selective search for object recognition.
\newblock {\em International journal of computer vision}, 104(2):154--171,
  2013.

\bibitem{girshick2015fast}
Ross Girshick.
\newblock Fast r-cnn.
\newblock In {\em Proceedings of the IEEE international conference on computer
  vision}, pages 1440--1448, 2015.

\bibitem{ren2015faster}
Shaoqing Ren, Kaiming He, Ross Girshick, and Jian Sun.
\newblock Faster r-cnn: Towards real-time object detection with region proposal
  networks.
\newblock In {\em Advances in neural information processing systems}, pages
  91--99, 2015.

\bibitem{liu2016ssd}
Wei Liu, Dragomir Anguelov, Dumitru Erhan, Christian Szegedy, Scott Reed,
  Cheng-Yang Fu, and Alexander~C Berg.
\newblock Ssd: Single shot multibox detector.
\newblock In {\em European conference on computer vision}, pages 21--37.
  Springer, 2016.

\bibitem{redmon2016you}
Joseph Redmon, Santosh Divvala, Ross Girshick, and Ali Farhadi.
\newblock You only look once: Unified, real-time object detection.
\newblock In {\em Proceedings of the IEEE conference on computer vision and
  pattern recognition}, pages 779--788, 2016.

\bibitem{lin2017focal}
Tsung-Yi Lin, Priya Goyal, Ross Girshick, Kaiming He, and Piotr Doll{\'a}r.
\newblock Focal loss for dense object detection.
\newblock In {\em Proceedings of the IEEE international conference on computer
  vision}, pages 2980--2988, 2017.

\bibitem{kang2017noscope}
Daniel Kang, John Emmons, Firas Abuzaid, Peter Bailis, and Matei Zaharia.
\newblock Noscope: optimizing neural network queries over video at scale.
\newblock {\em arXiv preprint arXiv:1703.02529}, 2017.

\bibitem{saha2016deep}
Suman Saha, Gurkirt Singh, Michael Sapienza, Philip~HS Torr, and Fabio
  Cuzzolin.
\newblock Deep learning for detecting multiple space-time action tubes in
  videos.
\newblock {\em arXiv preprint arXiv:1608.01529}, 2016.

\bibitem{xu2015learning}
Dan Xu, Elisa Ricci, Yan Yan, Jingkuan Song, and Nicu Sebe.
\newblock Learning deep representations of appearance and motion for anomalous
  event detection.
\newblock {\em arXiv preprint arXiv:1510.01553}, 2015.

\bibitem{ionescu2019object}
Radu~Tudor Ionescu, Fahad~Shahbaz Khan, Mariana-Iuliana Georgescu, and Ling
  Shao.
\newblock Object-centric auto-encoders and dummy anomalies for abnormal event
  detection in video.
\newblock In {\em Proceedings of the IEEE Conference on Computer Vision and
  Pattern Recognition}, pages 7842--7851, 2019.

\bibitem{cocsar2016toward}
Serhan Co{\c{s}}ar, Giuseppe Donatiello, Vania Bogorny, Carolina Garate,
  Luis~Otavio Alvares, and Fran{\c{c}}ois Br{\'e}mond.
\newblock Toward abnormal trajectory and event detection in video surveillance.
\newblock {\em IEEE Transactions on Circuits and Systems for Video Technology},
  27(3):683--695, 2016.

\bibitem{xu2019joint}
Huijuan Xu, Boyang Li, Vasili Ramanishka, Leonid Sigal, and Kate Saenko.
\newblock Joint event detection and description in continuous video streams.
\newblock In {\em 2019 IEEE Winter Conference on Applications of Computer
  Vision (WACV)}, pages 396--405. IEEE, 2019.

\bibitem{chau2011robust}
Duc~Phu Chau, Fran{\c{c}}ois Br{\'e}mond, Monique Thonnat, and Etienne
  Corv{\'e}e.
\newblock Robust mobile object tracking based on multiple feature similarity
  and trajectory filtering.
\newblock {\em arXiv preprint arXiv:1106.2695}, 2011.

\bibitem{stauffer1999adaptive}
Chris Stauffer and W~Eric~L Grimson.
\newblock Adaptive background mixture models for real-time tracking.
\newblock In {\em Proceedings. 1999 IEEE Computer Society Conference on
  Computer Vision and Pattern Recognition (Cat. No PR00149)}, volume~2, pages
  246--252. IEEE, 1999.

\end{thebibliography}
\bibliographystyle{unsrt}

%
%
%
%

\end{document}